\documentclass[sigconf]{acmart}
\usepackage{graphicx}
\usepackage{amsmath}
\usepackage{epsfig}
\usepackage{adjustbox}
\usepackage{dsfont}
\usepackage{float}
\usepackage{soul}
\usepackage{enumitem}
\usepackage{multirow}
\usepackage{booktabs}
\usepackage{pifont}
\usepackage{subcaption}
\usepackage{setspace}
\newcommand{\cmark}{\ding{51}}
\newcommand{\xmark}{\ding{55}}
\definecolor{Gray}{gray}{0.9}

\AtBeginDocument{%
  }

\copyrightyear{2022}
\acmYear{2022}
\setcopyright{rightsretained}
\acmConference[MM '22]{Proceedings of the 30th ACM International Conference on Multimedia}{October 10--14, 2022}{Lisboa, Portugal}
\acmBooktitle{Proceedings of the 30th ACM International Conference on Multimedia (MM '22), Oct. 10--14, 2022, Lisboa, Portugal}
\acmDOI{10.1145/3503161.3548027}
\acmISBN{978-1-4503-9203-7/22/10}

\makeatother

\begin{document}

\title{AVA-AVD: Audio-Visual Speaker Diarization in the Wild}

\author{Eric Zhongcong Xu}
\affiliation{
  \institution{Show Lab, National University of Singapore}
  \streetaddress{}
  \city{Singapore}
  \state{}
  \country{Singapore}
  \postcode{}
}
\email{zhongcongxu@u.nus.edu}

\author{Zeyang Song}
\affiliation{
  \institution{Show Lab, National University of Singapore}
  \streetaddress{}
  \city{Singapore}
  \state{}
  \country{Singapore}
  \postcode{}
}
\email{zeyang\_song@u.nus.edu}

\author{Satoshi Tsutsui}
\affiliation{
  \institution{Show Lab, National University of Singapore}
  \streetaddress{}
  \city{Singapore}
  \state{}
  \country{Singapore}
  \postcode{}
}
\email{satoshi@nus.edu.sg}

\author{Chao Feng}
\affiliation{
  \institution{Show Lab, National University of Singapore}
  \streetaddress{}
  \city{Singapore}
  \state{}
  \country{Singapore}
  \postcode{}
}
\email{fengchao336@gmail.com}

\author{Mang Ye}
\authornote{Corresponding author.}
\affiliation{
  \institution{Wuhan University}
  \streetaddress{}
  \city{Wuhan}
  \state{}
  \country{China}
  \postcode{}
}
\email{mangye16@gmail.com}

\author{Mike Zheng Shou}
\affiliation{
  \institution{Show Lab, National University of Singapore}
  \streetaddress{}
  \city{Singapore}
  \state{}
  \country{Singapore}
  \postcode{}
}
\email{mike.zheng.shou@gmail.com}

\renewcommand{\shortauthors}{Eric Zhongcong Xu et al.}

\begin{abstract}
Audio-visual speaker diarization aims at detecting ``who spoke when'' using both auditory and visual signals. Existing audio-visual diarization datasets are mainly focused on indoor environments like meeting rooms or news studios, which are quite different from in-the-wild videos in many scenarios such as movies, documentaries, and audience sitcoms. To develop diarization methods for these challenging videos, we create the AVA Audio-Visual Diarization (AVA-AVD) dataset. Our experiments demonstrate that adding AVA-AVD into training set can produce significantly better diarization models for in-the-wild videos despite that the data is relatively small. Moreover, this benchmark is challenging due to the diverse scenes, complicated acoustic conditions, and completely off-screen speakers. As a first step towards addressing the challenges, we design the Audio-Visual Relation Network (AVR-Net) which introduces a simple yet effective modality mask to capture discriminative information based on face visibility. Experiments show that our method not only can outperform state-of-the-art methods but is more robust as varying the ratio of off-screen speakers. Our data and code has been made publicly available at \textcolor{magenta}{\url{https://github.com/showlab/AVA-AVD}}.
\end{abstract}



\maketitle

\section{Introduction}\label{sec:intro}

\begin{figure}[t]
  \centering
  \includegraphics[width=0.88\linewidth]{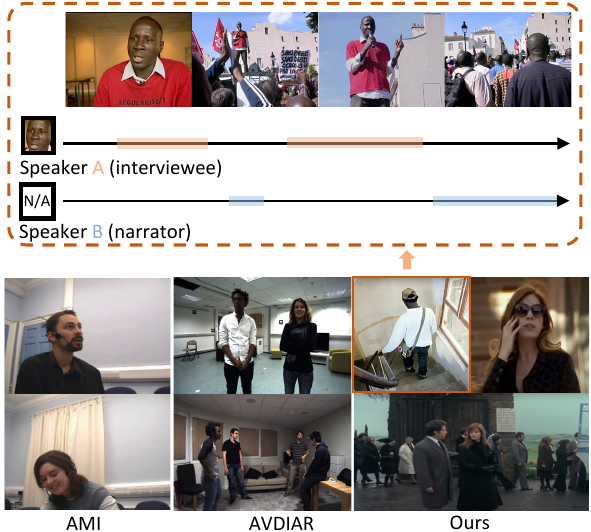}
  \caption{While existing audio-visual diarization datasets (AMI~\cite{carletta2005ami} and AVDIAR~\cite{gebru2017audio}) are recorded indoors, our AVA-AVD dataset includes in-the-wild scenarios and completely off-screen speakers, e.g., scene from a documentary film. When someone (Speaker A) is interviewed indoors, the video also shows him speaking in front of crowds of people outside. Meanwhile, a narrator (Speaker B) complements the context without showing her face at all. Diarizing this is challenging due to the noise (voices from crowds, traffic noise, etc.), which is not observed indoors and requires recognizing off-screen speakers of narrators.} 
  \label{fig:overview}
\end{figure}

Speech effectively delivers rich information verbally and often visually. A documentary film (Figure~\ref{fig:overview}), for example, features a person's life using his interview, his speech in public, and complementary comments by a narrator without showing her face. Organizing the rich information embedded in speech in videos (meeting videos, interviews, news broadcasts, documentaries, comedy shows, etc.) is necessary for many practical applications such as  translation~\cite{Afouras18b},
summarization~\cite{rochan2019video}, social analysis~\cite{grauman2022ego4d,yan2020social}, etc. 

The first step for developing these applications is diarization, recognizing ``who spoke when.'' It is a task to identify speakers and localize the start and end times of each person's utterances. While early work on speaker diarization is based solely on the audio stream~\cite{sell2018diarization,wang2018speaker,zhang2019fully}, recent studies~\cite{chung2019said, chung2020spot,gebru2017audio} attempt to exploit both audio and visual cues. These methods improve diarization performance by combining similarity scores of two independent modalities or incorporating synchronization between utterances and lip motion. However, existing audio-visual diarization datasets~\cite{carletta2005ami, chung2020spot, gebru2017audio} are mainly collected from indoor meetings, chats, or news. Therefore, the benchmark built on these datasets cannot reflect the system's performance in {\it unseen} scenarios or \textit{in-the-wild} videos (e.g, movies, documentaries, and audience sitcoms).  

In this paper, we develop a new dataset focusing on the following two aspects of in-the-wild videos. (1) {\bf Scene diversity}: Unlike indoor videos, in-the-wild videos have various acoustic environments (e.g., traffic noise, background music, etc.) and complex editing structures (e.g., frequent scene changes). (2) {\bf Completely off-screen speakers}: In-the-wild videos often have completely off-screen speakers, such as speakers on the phone or radio, and narrators in the background. These speakers never appear in the field-of-view and thus audio-visual correspondence is not observed. Unfortunately, diarization systems developed on indoor datasets are difficult to generalize well for these in-the-wild scenarios due to the large domain gap. The first roadblock to remedy this issue is the lack of a proper testbed i.e. an in-the-wild audio-visual diarization benchmark. To this end, we introduce a challenging audio-visual diarization dataset i.e. AVA-Audio-Visual Diarization (AVA-AVD). The video sources of this dataset are the AVA-Active Speaker~\cite{roth2020ava}. We select 117 unique videos that have high resolution and no dub. As shown in Figure~\ref{fig:overview}, AVA-AVD has various complicated scenarios with a wide spectrum of daily activities, it provides both a challenging test set and diverse training data.

In addition, to address the challenges (e.g., off-screen speakers) in audio-visual diarization, we propose a simple yet effective Audio-Visual Relation Network (AVR-Net) which explores the cross-modal links between the speaker's face and voice and handle the invisible speakers. Our key insights are: (1) the consistency between the face and voice~\cite{nagrani2018learnable,nagrani2018seeing,gao2021visualvoice} is a useful cue for speaker diarization. For example, a female speaker may have a higher voice frequency; a heavier person may have a wide range of sound intensities~\cite{gao2021visualvoice,barsties2013body}. Humans can by nature capture such cross-modal 
association and then separate and group the speakers, while the commonly used late fusion strategy ignores these relationships. (2) Depending on the face visibility of two candidate speakers, we have four cases, i.e. {\it A} vs. {\it A}, {\it A} vs. {\it A-V}, {\it A-V} vs. {\it A}, and {\it A-V} vs. {\it A-V}, where {\it A} denotes audio, and {\it A-V} denotes audio-visual. Therefore, we introduce modality mask to process these cases in a unified model. Furthermore, instead of learning discriminative embedding vectors and measuring distances linearly~\cite{nagrani2018seeing,nagrani2018learnable,sari2021multi,wen2018disjoint,qian2021audio}, our model utilizes Siamese~\cite{koch2015siamese} architecture and the distance metric is also optimized during training.

In summary, our main contributions are as follows:
\begin{itemize}
\item We introduce a new challenging audio-visual diarization dataset AVA-AVD, and experimentally show that SOTA methods perform poorly on such in-the-wild videos.  
\item Besides the test set, we also provide a good amount of valuable training data, which can decrease the diarization error rate by 17.1\%.
\item We propose a simple yet effective baseline model Audio-Visual Relation Network (AVR-Net) which can handle missing modalities using learnable modality masks. 
Comparing to SOTA methods, our model improves diarization performance with a relative gain of 6.6\%, and is more robust as varying the ratio of off-screen speakers.
\end{itemize}

\section{Related Work}
\noindent{\bf Audio-only Speaker Diarization.} 
Speaker diarization using only audio streams has been studied extensively in recent years. Most diarization systems~\cite{ryant2019second,xiao2021microsoft,landini2022bayesian,wang2018speaker} employ a multi-stage framework, including voice activity detection~\cite{johnston2012webrtc}, embedding extraction~\cite{kenny2010bayesian,dehak2010front,snyder2018x,wan2018generalized}, and clustering~\cite{day1984efficient,ng2002spectral}. Each component of the system will affect the final diarization performance. A typical audio-only diarization system adopts off-the-shelf voice activity detection and speaker verification models. Therefore, prior works about audio-only diarization focused on denoising~\cite{ryant2019second}, clustering algorithm~\cite{diez2020optimizing}, and handling overlap speech~\cite{landini2021analysis}. A recent work~\cite{landini2022bayesian} adopts Bayesian clustering. Although it achieves state-of-the-art performance on the standard audio-only diarization benchmark~\cite{carletta2005ami}, the system cannot generalize well for in-the-wild scenarios with complicated acoustic conditions. 

Another promising direction is end-to-end diarization. It has exhibited several advantages, such as a simpler pipeline and the ability to cope with speech overlap. However, existing end-to-end methods can only handle a limited number of speakers~\cite{medennikov2020target,Horiguchi2020} and rely heavily on synthetic data for training~\cite{huang2020speaker}. At the current stage, the performance of end-to-end systems still cannot outperform their multi-stage counterparts.

\noindent{\bf Audio-Visual Speaker Diarization.} 
Facial attributes and lip motion are highly related to speech~\cite{yehia1998quantitative}. In particular, visual information can help handle speech overlap and predict both active speakers and speaker changes. Therefore, several methods that leverage audio and visual cues for diarization are motivated by the synergy between utterances and lip movements. These methods adopt techniques such as mutual information~\cite{hershey1999audio,noulas2011multimodal}, canonical correlation analysis~\cite{hotelling1992relations}, and deep learning~\cite{Chung16a,chung2019perfect,chung2019said}. In addition, audio-visual correspondence is also used for associating talking faces and voice tracks~\cite{tao2021someone,chung2020spot,xia2020online}. Based on the paired audio-visual inputs, different modalities are fused by linear combination~\cite{chung2020spot,chung2019said}, temporal alignment~\cite{el2014audiovisual,kapsouras2017multimodal}, or Bayesian method~\cite{gebru2017audio}. A recent work~\cite{chung2019said} proposes an audio-visual speaker diarization system using ``enroll first, diarize later" paradigm. Compared with audio-only diarization, the combination of auditory and visual information improves the performance by a large margin, yet this paradigm is difficult to generalize in realistic scenarios because (1) it is hard to collect ground truth profiles for each person or character in unseen videos; (2) speakers outside the field-of-view cannot be enrolled due to the lack of audio-visual correspondence.

In contrast, our system does not require ground truth face images for enrollment. It takes both utterances and associated faces as input and can learn to compute the similarity of two speakers. Our model learns both uni-modal and cross-modal links of two candidate speakers. Instead of using linear metrics such as cosine similarity or Euclidean distance, the measurement function in our model is also optimized during training.

\noindent{\bf Speaker Verification with Faces and Voices.}
Off-the-shelf verification methods including speaker verification~\cite{wan2018generalized,chung2020in} and face recognition~\cite{deng2019arcface,parkhi2015deep} are often used to extract discriminative embeddings. These models are pretrained on large-scale datasets~\cite{Chung18b,Nagrani17,cao2018vggface2,guo2016ms} with a large number of identities. Thus, they can generalize well for in-the-wild videos. Based on the pretrained models, many interesting downstream tasks have been explored, such as deep clustering~\cite{jiang2019dm2c,hu2017learning}, face clustering~\cite{tapaswi2019video,jin2017end}, person search~\cite{xia2020online}, person clustering~\cite{brown2021face}, as well as speaker diarization~\cite{landini2022bayesian,chung2019said}. Although the above verification models have a relatively strong generalization ability, they are all designed for uni-modal verification or identification tasks. However, cross-modal links between one person's face and voice can also reveal the speaker's identity and have been extensively studied in~\cite{nagrani2018learnable,nagrani2018seeing,gao2021visualvoice,wen2018disjoint,wen2021seeking}. 

To this end, our model also leverages this cross-modal correlation to recognize the speaker's identity. We design a simple yet effective modality mask to dynamically learn both uni-modal and cross-modal similarities in a unified framework.

\begin{table*}[t]
\centering
\small
\caption{Comparisons with existing audio-visual speaker diarization datasets. available denotes whether the dataset is publicly accessible or not. {\bf os speakers} denote completely off-screen speakers. Numbers separated by $/$ denote minimum, average, and maximum.}
\begin{tabular}{lccccccccc}
\hline
{\bf Dataset}&{\bf available}&{\bf scenario}&{\bf \#videos} &{\bf duration} &{\bf \#IDs} &{\bf \#speakers}&{\bf \#os speakers} &{\bf speech\%} &{\bf language}\\
\hline
AMI~\cite{carletta2005ami} Corpus &A\cmark V\cmark &meetings &684 &100h &189 &3/4.0/5 &0/0/0 &50.5/83.2/96.4 &En\\
AVDIAR~\cite{gebru2017audio} &A\cmark V\cmark &chat &27 &21m &11 &1/2.2/4 &0/0/0 &23.9/82.6/96.2 &En,Fr\\
VoxConverse~\cite{chung2020spot} &A\cmark V\xmark&debate, news&448 &63h50m &- &1/5.6/21 &- &10.7/91.3/100.0 &En\\
AVA-AVD (Ours) &A\cmark V\cmark&diverse daily activities &351 &29h15m &1500 &2/7.7/24 &0/2.4/20 &12.0/46.0/84.3 &Multi(>6)\\
\hline
\end{tabular}
\label{tab:stat}
\end{table*}

\section{AVA-AVD}
In this section, we describe the video source (Sec.~\ref{source}), our annotation process (Sec.~\ref{annotation}), and the statistics (Sec.~\ref{statistics}) of AVA Audio-Visual Diarization (AVA-AVD) dataset.

\subsection{Video Source}
\label{source}
Following the recent studies~\cite{alcazar2021apes,brown2021face}, instead of collecting or recording new videos, we build AVA-AVD dataset upon a publicly available dataset, i.e. AVA-Active Speaker\cite{roth2020ava}.

\noindent{\bf Motivations.} There are three main reasons: (1) the diverse acoustic and visual scenarios in~\cite{roth2020ava} are challenging for audio-visual diarization task; (2) active speaker detection is a critical prior step for audio-visual diarization, by adding extra annotations, we can enrich this dataset and set up a complete benchmark; (3) with the ground truth face labels and active speaker labels provided by ~\cite{roth2020ava}, we can stay focused on the components which are closely related to diarization task itself (similar to the protocol introduced in~\cite{sell2018diarization}).

\noindent{\bf Original annotations.} AVA-Active Speaker has 144 videos in total. This collection features movies of different races, ample dialogues, and diverse languages and genres. It also has labels for faces and active speakers, which can aid the association between utterances and faces. We select 117 high-quality videos without any dub because: (1) audio-visual synergy is destroyed in dubbed videos; (2) it could bring ambiguity if the same person dubs for multiple characters. In each video, a 15-minute clip is annotated by the AVA-Active Speaker. Note that the time complexity for verifying speaker ID is $O(N^2)$. A too-short video clip is less challenging and may not have enough speakers, while a too-long clip will increase the speaker matching difficulty and time consumption, which may jeopardize human annotators' confidence. Therefore, we carefully selected the clip length by trial and error, a 15-minute clip is further divided into three 5-minute clips.
\begin{figure}[h]
  \centering
  \begin{subfigure}{0.42\linewidth}
    \centering
    \includegraphics[width=\linewidth]{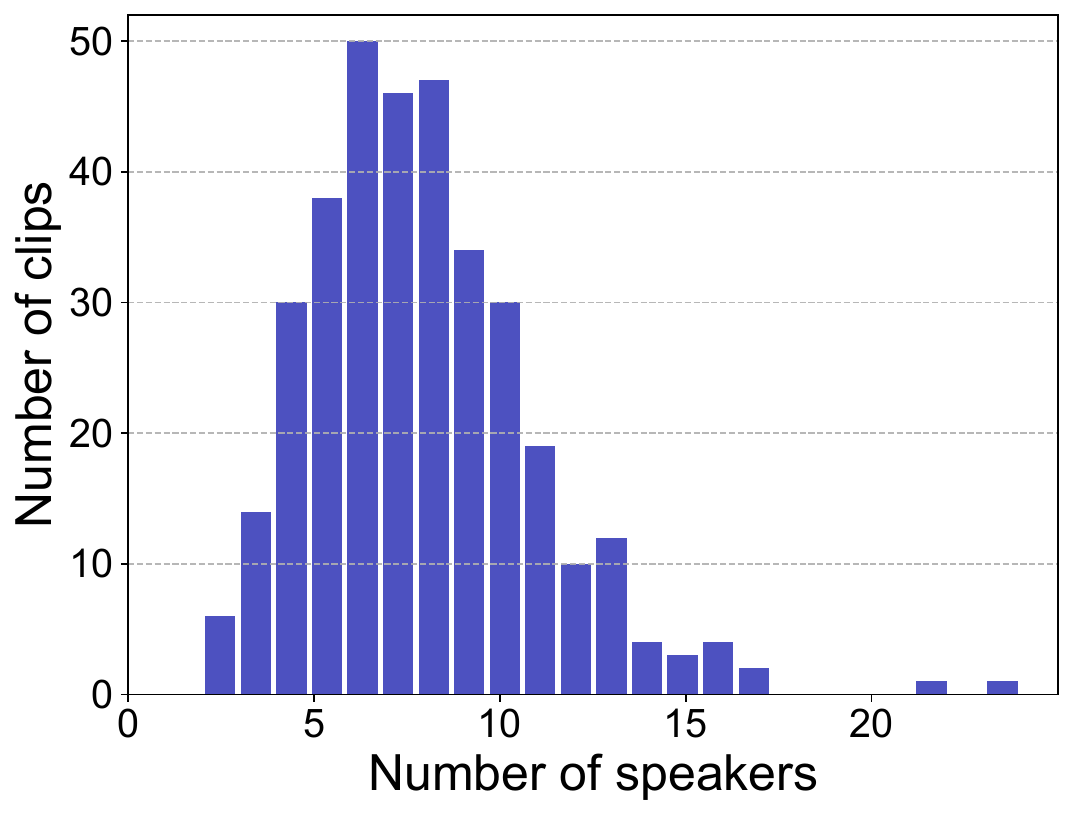}
    \caption{}
    \label{fig:stat:a}
  \end{subfigure}
  \begin{subfigure}{0.425\linewidth}
    \centering
    \includegraphics[width=\linewidth]{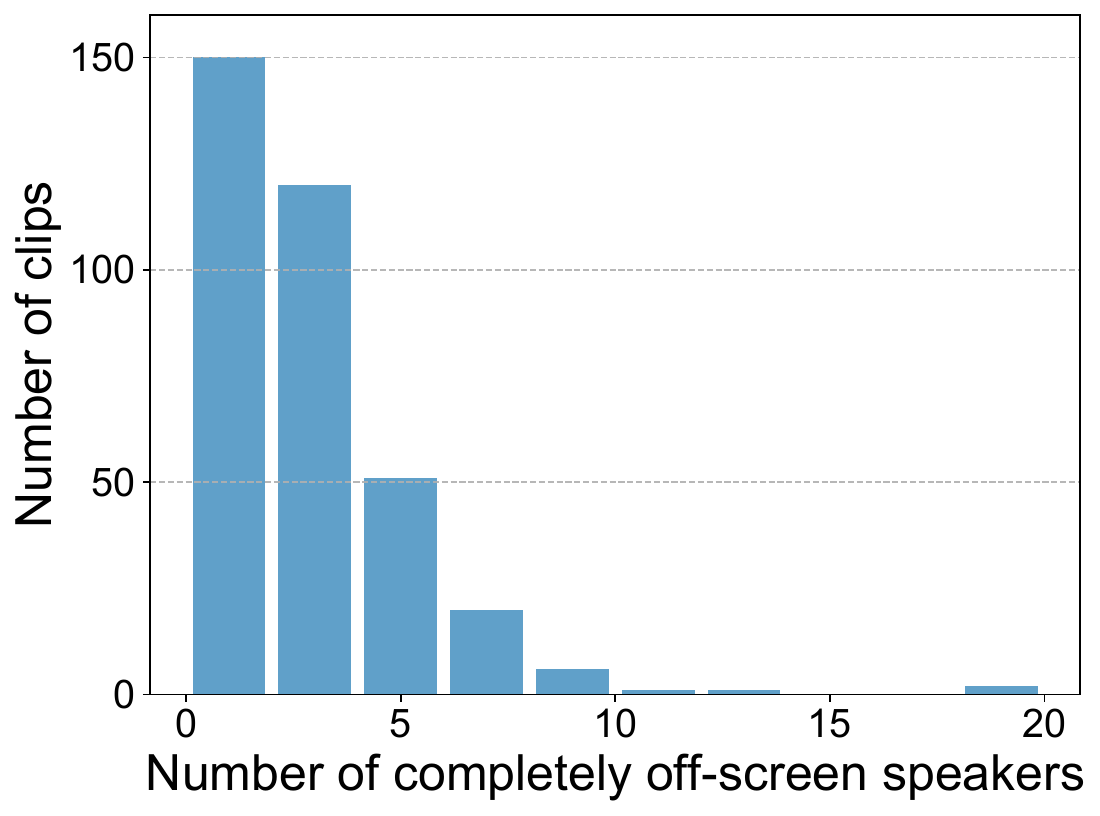}
    \caption{}
    \label{fig:stat:b}
  \end{subfigure}
  \begin{subfigure}{0.43\linewidth}
    \centering
    \includegraphics[width=\linewidth]{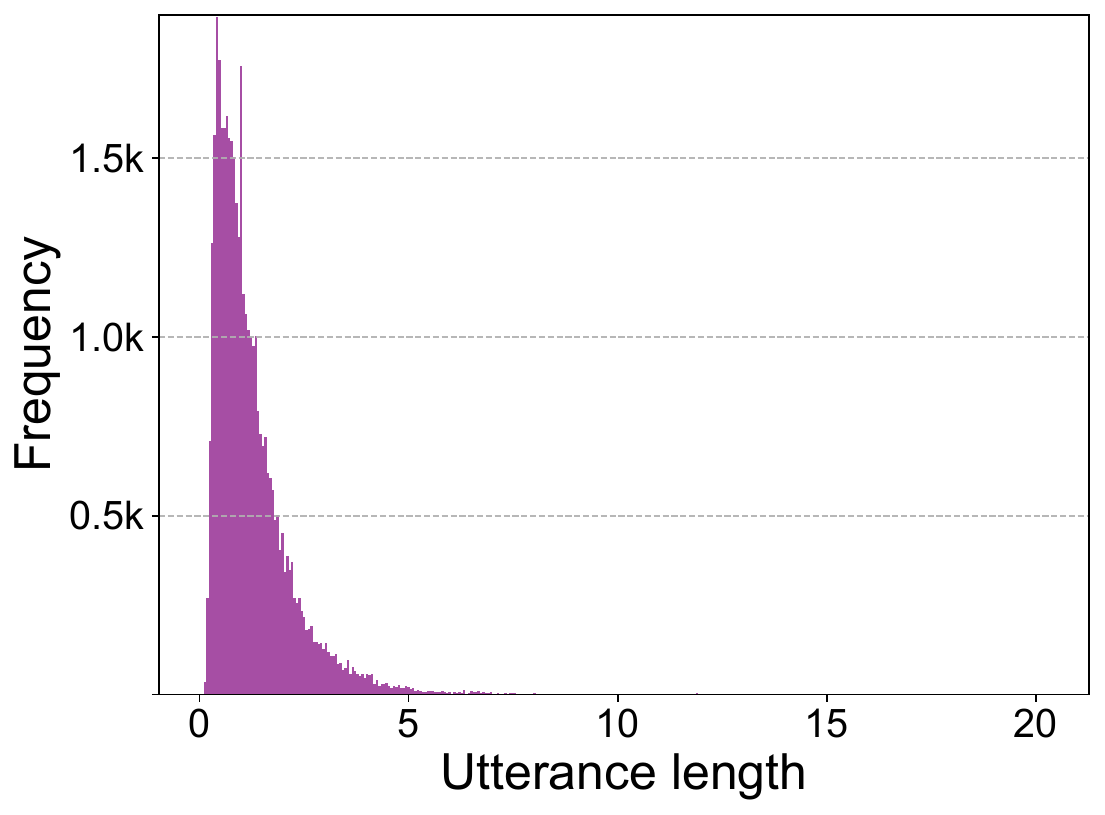}
    \caption{}
    \label{fig:stat:c}
  \end{subfigure}
  \begin{subfigure}{0.425\linewidth}
    \centering
    \includegraphics[width=\linewidth]{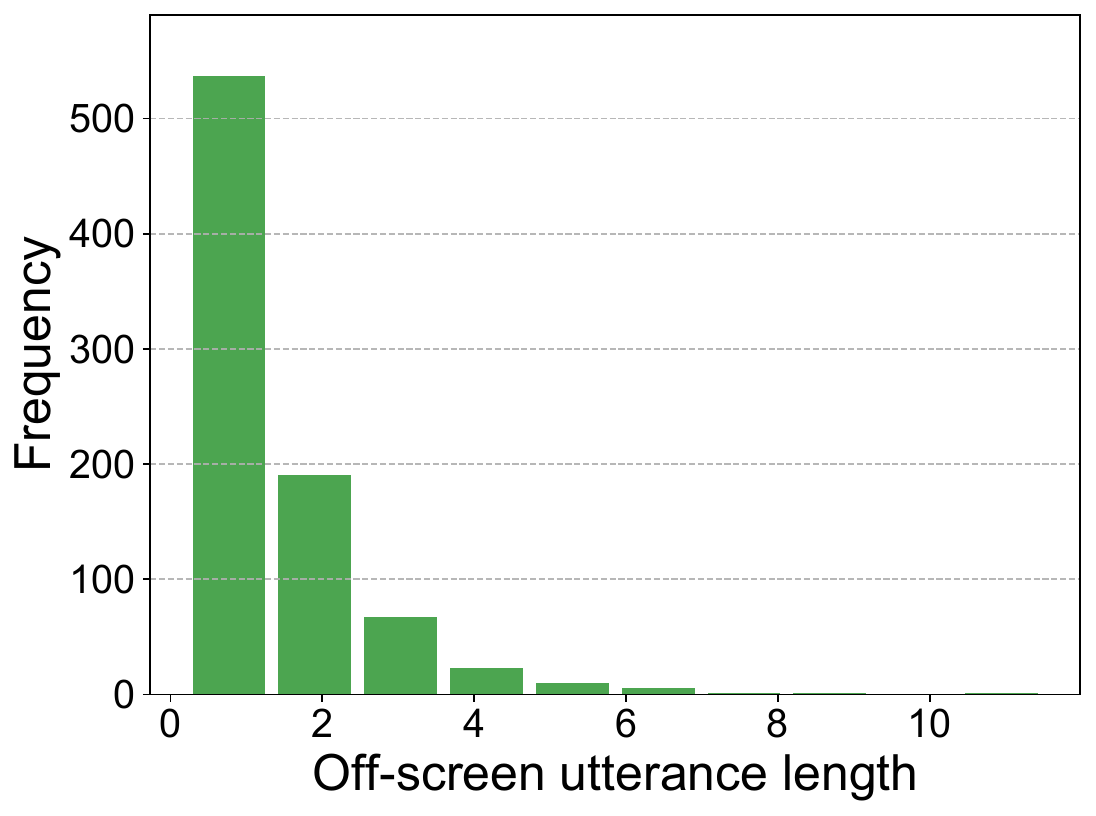}
    \caption{}
    \label{fig:stat:d}
  \end{subfigure}
  \caption{Histograms for AVA-AVD statistics showing: (a) number of clips {\it vs.} number of speakers in each clip; (b) number of clips {\it vs.} number of completely off-screen speakers in each clip; (c) duration of speech segments; (d) duration of speech segments from off-screen speakers.
}
  \label{fig:stat}
\end{figure}
\subsection{Annotation Process}
\label{annotation}
Each video clip is annotated through the following steps:

\noindent{\bf Generating face and voice tracks.} Annotating faces frame by frame and labeling speech segments are time-consuming. Since AVA-Active Speaker provides annotated face crop tracks, we keep face tracks that contain audible speeches. Furthermore, the proposals for rough speech segments are automatically generated using the timestamps of speaking face tracks.

\noindent{\bf Labeling face tracks.} Given the automatically generated face tracks, annotators need to identify their speaker ID. Human annotators watch the video and listen to the concurrent audio stream to assign video-level identity labels to each face. Details of our annotation tools can be found in {\it Sup Mat}.

\noindent{\bf Adjusting speech segments.} Some speakers are temporally invisible in the video frame due to the rapid motion or video editing structure, etc. Thus, human annotators need to refine the onset and offset of the speech segment proposals and get accurate segment labels. In addition, neighboring short segments produced by the same speaker are merged.

\noindent{\bf Labeling off-screen speakers.} Another challenging feature of in-the-wild videos is that there may exist completely off-screen speakers, e.g., speaker talking on the phone, broadcasting on the radio, or narration in the background. In this stage, the annotators thoroughly watch and listen to the video clips and annotate the IDs for the utterances generated by the off-screen speakers. Relying only on the voice similarities, annotators label each utterance with either an existing speaker ID or a new one.

\noindent{\bf Mutual verification.} At last, to guarantee the label quality, the annotations are manually verified by different annotators. The average time cost for completing the above steps is 8 times of the video duration.

\begin{figure*}[ht]
  \centering
  \includegraphics[width=0.8\linewidth]{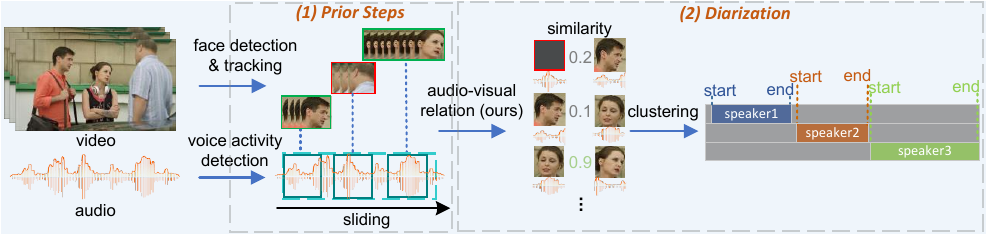}
  \caption{Overview of our audio-visual diarization system. Faces highlighted in red boxes are invisible, but our system can process this case in a unified framework.
  Our model directly predicts similarity score for two candidate speakers.} 
  \label{fig:sys}
\end{figure*}

\subsection{Statistics}
\label{statistics}
{\bf Comparisons with existing datasets.} Table~\ref{tab:stat} summarizes statistics of existing audio-visual diarization datasets. Compared to existing datasets that only have indoor activities such as meetings, free chats, and debates, AVA-AVD has a wide spectrum of activities. It provides around 1500 unique identities with voice tracks and/or faces. We compute the number of identities by mapping the face tracks into global labels based on the annotations released by~\cite{alcazar2021apes}.
Figure~\ref{fig:stat:a} and~\ref{fig:stat:c} illustrate the number of speakers and the long-tail utterance length distribution. Especially, the statistics for the off-screen speakers are shown in Figure~\ref{fig:stat:b} and~\ref{fig:stat:d}. Also, unlike AMI and VoxConverse, our AVA-AVD includes videos in multiple languages such as English, French, Chinese, German, Korean, Spanish, etc.

Finally, we split 243 clips (20.25 hours) for training, 54 clips (4.5 hours) for validation, and 54 clips (4.5 hours) for testing. More details of the dataset partition strategy can be found in {\it Sup Mat}. 
\section{Method}
Our audio-visual diarization system is shown in Figure~\ref{fig:sys}. We first formulate the diarization problem (Sec.~\ref{pdef}); then we present the audio-visual relation network and introduce how to learn audio-visual similarity with missing faces (Sec.~\ref{AVR-Net}). 

\subsection{Problem Definition}
\label{pdef}
\noindent {\bf Prior steps.} Given a video with multiple speakers (Figure~\ref{fig:sys}), our goal is to localize the audible utterances and label them with video level speaker identity leveraging audio-visual cues. Following a commonly adopted multi-stage paradigm~\cite{chung2020spot,chung2019said,wang2018speaker}, we can either run face detection and tracking~\cite{zhang2017s3fd}, voice activity detection\cite{landini2021analysis}, and active speaker detection~\cite{tao2021someone} in advance or directly use the ground truth results similar to~\cite{landini2022bayesian}. Then, we get a set of speech segments $A_i^{[s_i,e_i]}$, where $i \in \{1,2...N\}$, $s_i$ and $e_i$ are onset and offset. We also have the talking faces $I_t^k$ at each time $t$, where $t \in \{1,2...T\}$, $k\in \{1,2...K_t\}$, and $K_t$ is the maximum number of speakers at $t$. For each speech segment, we ensemble an utterance $A_i^{[s_i,e_i]}$ and one image $I_i$ sampled from track $\{I_t^k$, $t\in[s_i,e_i]\}$ into an audio-visual pair. 

\noindent {\bf Diarization.} Our goal is to train a network $F_{\phi}$, which computes the similarities between two audio-visual pairs
\begin{equation}
s_{i,j}=F_{\phi}((A_i^{[s_i,e_i]}, I_i), (A_j^{[s_j,e_j]}, I_j)),
\end{equation}
where $i,j \in \{1,2...N\}$. Then we get the similarity graph represented by the adjacent matrix $S^{N\times N}$. Based on the graph $S$, we perform agglomerative hierarchical clustering~\cite{day1984efficient} following~\cite{chung2020spot,chung2019said,landini2022bayesian} and finally get the diarization results. Previous works measure similarities by cosine similarity or Euclidean distance based on the embedding vectors extracted by off-the-shelf models, while we integrate these two separated steps into one model.

\subsection{Audio-Visual Relation Network}
\label{AVR-Net}
\noindent{\bf Backbone.} Pretrained on large-scale datasets, speaker verification and face recognition models have strong generalization ability and are applied directly in various downstream tasks~\cite{jiang2019dm2c,hu2017learning,tapaswi2019video,jin2017end,brown2021face,xia2020online,landini2022bayesian,chung2019said}. We follow these works and utilize the backbones of off-the-shelf models~\cite{chung2020in, deng2019arcface} to extract intermediate voice and face features respectively.

\begin{figure}[t]
\centering
\includegraphics[width=0.45\textwidth]{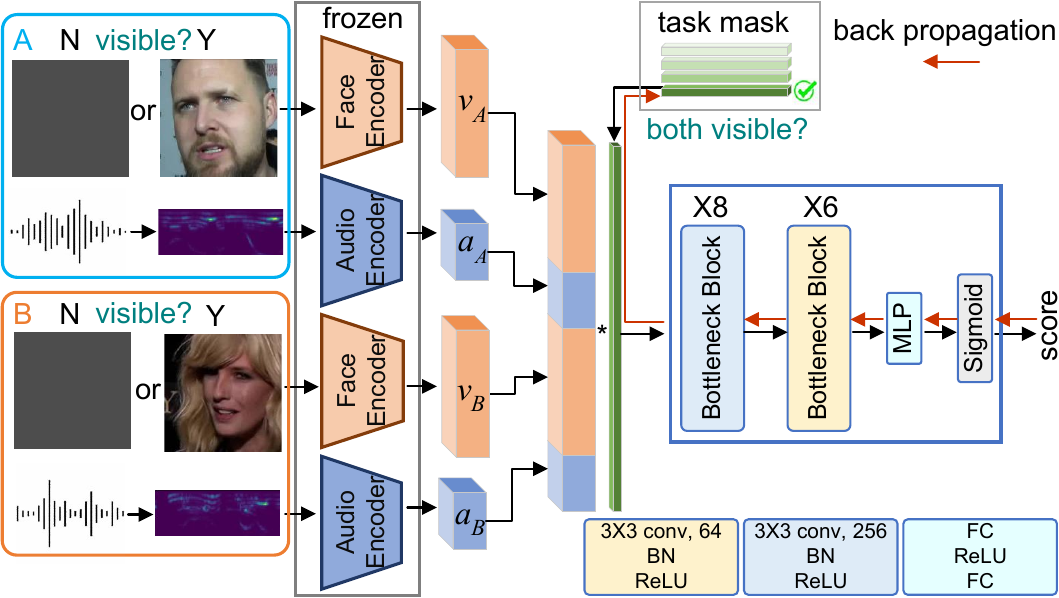}
\caption{Architecture of our Audio-Visual Relation Network. We use pretrained backbones to extract voice and face features. The intermediate features are concatenated and multiplied with a learnable modality mask, which is selected based on the visibility of the face. The relation module predicts the similarity score of two candidate speakers directly.} 
\label{fig:AVR-Net}
\end{figure}

\noindent{\bf Relation module.} As shown in Figure~\ref{fig:AVR-Net}, given two audio-visual pairs $(A_i^{[s_i,e_i]}, I_i), (A_j^{[s_j,e_j]}, I_j)$, the audio and visual backbones encode them into intermediate feature pairs: $F_A\in \mathbb{R}^{C_A\times H_A\times W_A}$ and $F_I\in \mathbb{R}^{C_I\times H_I\times W_I}$. We first apply adaptive pooling on $F_A$ and then concatenate two modalities along the channel axis. Therefore, we get the feature for each audio-visual pair $F_{P}\in\mathbb{R}^{(C_I+C_A)\times H_I\times W_I}$. Then we further concatenate two audio-visual pairs and feed them into the relation module. In audio-visual diarization, it is common to encounter off-screen speakers due to low resolution, sharp motion, light conditions, head pose, or film editing. Depending on the face visibility, we distinguish the comparison cases. Specifically, here we have four cases: {\it A} vs. {\it A}, {\it A} vs. {\it A-V}, {\it A-V} vs. {\it A}, and {\it A-V} vs. {\it A-V}. To handle different cases within a single network, we draw inspiration from ~\cite{carion2020end,sun2021sparse} and design learnable modality masks $M_i=\mathbb{R}_i^{(C_I+C_A)*2}$, $i\in\{0,1,2,3\}$. Prior to the relation module, we compute interactions between audio-visual features and the modality mask by channel-wise multiplication. With the modality masks, our model can dynamically activate the input features (discussed in Sec.~\ref{exp:analysis}). Our relation module consists of two stacked ResNet-style layers. Finally, a simple MLP layer maps the output into one dimension and then predicts the similarity score using Sigmoid.

\noindent{\bf Loss function.} AVR-Net is trained by
\begin{equation}
\phi \leftarrow \underset{\phi}{\operatorname{argmin}} \sum_{i=1}^{N} \sum_{j=1}^{N}\left(s_{i, j}-\mathbf{1}\left(y_{i}==y_{j}\right)\right)^{2}
\label{obj},
\end{equation}
where $N$ is the batch size, $y_{i}$ and $y_{j}$ denote the identity labels. The goal of our AVR-Net is regressing 1 if two pairs are from the sample person and 0 vice versa.
Similar to~\cite{sung2018learning}, we adopt mean square error (MSE) loss because the similarity score is continuous and can be considered a regression problem.

\noindent{\bf Training.} Diarization datasets only provide annotations in Rich Transcription Time Marked (rttm) format. Several end-to-end diarization systems~\cite{medennikov2020target,Horiguchi2020,huang2020speaker} are trained on synthesized datasets due to the limited dataset scales, while multi-stage works~\cite{ryant2019second,xiao2021microsoft,landini2022bayesian,wang2018speaker} never train verification models on the diarization dataset. The only parameter they need to optimize is the threshold for clustering.

In contrast, we explore training the audio-visual relation network on both speaker verification datasets~\cite{Chung18b,Nagrani17} and our AVA-AVD diarization dataset. One of the challenges of diarization datasets is that they only provide video-level speaker identities instead of global ones. There are two solutions to utilize the local identities and at the same time mitigate the drawbacks: (1) {\it treat the video level identity labels as global ones:} It has been studied in unsupervised re-identification works~\cite{ge2020selfpaced} that splitting the same instance into different instances will not degrade performance too much; (2) {\it sample positive data within one video and sample negative data across videos:} This sampling strategy can avoid ambiguous labels while largely keep the diversity of negative samples. Experimentally, we find the second solution more effective in training our AVR-Net.

In addition, previous methods finetune a threshold on the validation set and fix this parameter during inference. In this work, we propose to automatically search the parameters during training and select the best model according to validation performance rather than evaluation metrics of pretraining (i.e. speaker verification or recognition) tasks. 

\noindent{\bf Inference.} During inference, we use a common sliding window approach to process the input sequence segment by segment for videos of arbitrary length. We randomly select one image from the face tracks and use the threshold tuned on validation set.

\section{Implementation Details}
Our audio-visual relation network is implemented in PyTorch. For all of the experiments, we resize the input faces to $112\times112$. Following a standard preprocessing step~\cite{wen2018disjoint,deng2019arcface}, we run face alignment using RetinaFace~\cite{Deng2020CVPR}. During training, to synthesize the missing cases, we replace the visible face with zero under a missing probability of 0.5.
The input speech segments are 2s and sub-sampled to 16 kHz. The spectrogram is extracted every 10ms with a window of 25ms on the fly. 
Our network is optimized with Adam using a learning rate of 5$\times$10$^{-4}$. We train the model for 80000 iterations on a single RTX 3090 GPU.

\section{Experiments}
In this section, we first run the state-of-the-art systems to study the challenges of AVA-AVD (Sec.~\ref{exp:challenge}). Second, we investigate the effectiveness of AVA-AVD training data (Sec.~\ref{exp:train}). Then we compare AVR-Net with state-of-the-art methods (Sec.~\ref{exp:compare}) and visualize the qualitative results (Sec.~\ref{exp:qual}). Finally, the proposed modality masks are studied in Sec.~\ref{exp:analysis}.

\subsection{Datasets}
{\bf VoxCeleb~\cite{Nagrani17}} and {\bf VoxCeleb2~\cite{Chung18b}} contains Youtube interview videos for 1,251 and 6,112 celebrities respectively. In addition to AVA-AVD, we also use these two datasets in experiments.

\subsection{Evaluation Metric}
The diarization performance is evaluated by Diarization Error Rate (DER), which is lower the better. It contains three terms: Missing Detection (MS), False Alarm (FA), and Speaker Error (SPKE). We use the evaluation tools released by~\cite{istrate2005nist} and adopt the {\bf Fair} protocol defined in~\cite{landini2022bayesian}, which considers overlap speech and includes an acceptance margin of 0.25s.

\subsection{Evaluate the Challenges of AVA-AVD}
\label{exp:challenge}
To evaluate the challenging characteristics of our AVA-AVD dataset, we first run the state-of-the-art audio-visual (WST~\cite{chung2019said}) and audio-only (VBx~\cite{landini2022bayesian}) diarization systems.

\begin{table}[t]
\centering
\small
\caption{The performance of existing diarization systems. DER increases a lot on our AVA-AVD, which implies the challenges of our dataset. GT: oracle VAD derived from diarization labels. W: WebRTC\cite{johnston2012webrtc} VAD algorithm. E: a simple energy-based VAD method introduced in~\cite{landini2021analysis}. We report results with energy-based VAD because it yields slightly better results. WST method is reproduced by ourselves. $\dag$ denotes the results directly cited from \cite{chung2019said}.}
\begin{tabular}{lccccccc}
\hline
&Test set &VAD & MS   & FA & SPKE & DER\\
\hline
\multirow{6}{*}{WST\cite{chung2019said}}&\multirow{2}{*}{AMI-ES$\dag$}  &GT     &5.6 &0.0&4.8&10.4 \\
& &W &10.5  &6.6 &4.0&21.1 \\
&\multirow{2}{*}{AMI-IS$\dag$}  &GT     &6.5 &0.0&7.1&13.6 \\
& &W   &11.2  &4.0 &6.2&21.3 \\
&\multirow{2}{*}{AVA-AVD} &GT   &3.08  &0.0 &38.95 &42.04 \\
& &E   &11.63 &40.63 &36.15 &88.41 \\
\hline
&\multirow{2}{*}{AMI} &GT   &9.55  &0.0 &2.98 &12.53 \\
VBx\cite{landini2022bayesian}&&E   &30.28  &1.05 &7.32 &38.65 \\
(ResNet101)&\multirow{2}{*}{AVA-AVD} &GT   &2.92  &0.0 &18.45 &21.37 \\
& &E   &8.71  &44.59 &17.59 &70.88\\
\hline
\end{tabular}
\label{tab:test}
\end{table}

\noindent{\bf Audio-Visual diarization system.} The WST system achieves 10.4 on AMI-ES subset, and 13.6 on AMI-IS subset. It enrolls the audio segments of each speaker based on the ground truth speaker profiles using synchronization model~\cite{Chung16a}. However, this enrollment process would fail for the {\it completely off-screen speakers} due to the lack of visual inputs. Therefore, WST only achieves a DER of 42.04 on AVA-AVD. The large speaker error (38.95) mainly comes from the {\it enrollment failures}. Noted that we use the open-source pretrained SyncNet\cite{Chung16a} instead of~\cite{chung2019perfect}.

\noindent{\bf Audio-only diarization system.} VBx is an audio-only diarization system. We use the same {\bf ResNet101} model released by the authors. Unlike WST, it uses clustering instead of enrollment. Completely off-screen speakers will not cause any system-level failure for VBx, thus it outperforms WST by 49.17\% on AVA-AVD. However, compared with AMI, VBx has a much higher SPKE (18.45) and DER (21.37) on AVA-AVD because AVA-AVD has multiple languages and diverse scenarios.

\noindent{\bf Challenging acoustic environments.} To evaluate the acoustic conditions in our AVA-AVD, we also investigate the effects of voice activity detection (VAD). For all of the experiments, we report the results using energy-based VAD and oracle VAD respectively. In general, the false alarm of both WST (40.63) and VBx (44.59) on AVA-AVD is large because AVA-AVD has a large amount of background music, traffic noise, laughter, etc. WST is slightly better due to the enrollment mechanism, but its false alarm (40.63) on AVA-AVD is significantly higher than on AMI because the latter is only recorded in silent meeting rooms. Different from other datasets with only indoor scenarios, simple energy-based VAD systems cannot work well on AVA-AVD. Other VAD methods such as combining VAD with ASR deserve further exploration. In the following experiments, to study the performance of AVR-Net, we report results using {\it oracle VAD} and {\it ground truth active speaker labels} unless stated.
\subsection{Necessity of AVA-AVD Train Set}
\label{exp:train}
Existing audio-visual diarization datasets~\cite{carletta2005ami,gebru2017audio} have less speaker diversity and thus no data is partitioned for training. However, since our AVA-AVD dataset has a large number of identities, we also split a train set and investigate its effectiveness in in-the-wild scenarios.
\begin{figure}[h!]
  \centering  
  \includegraphics[width=0.8\linewidth]{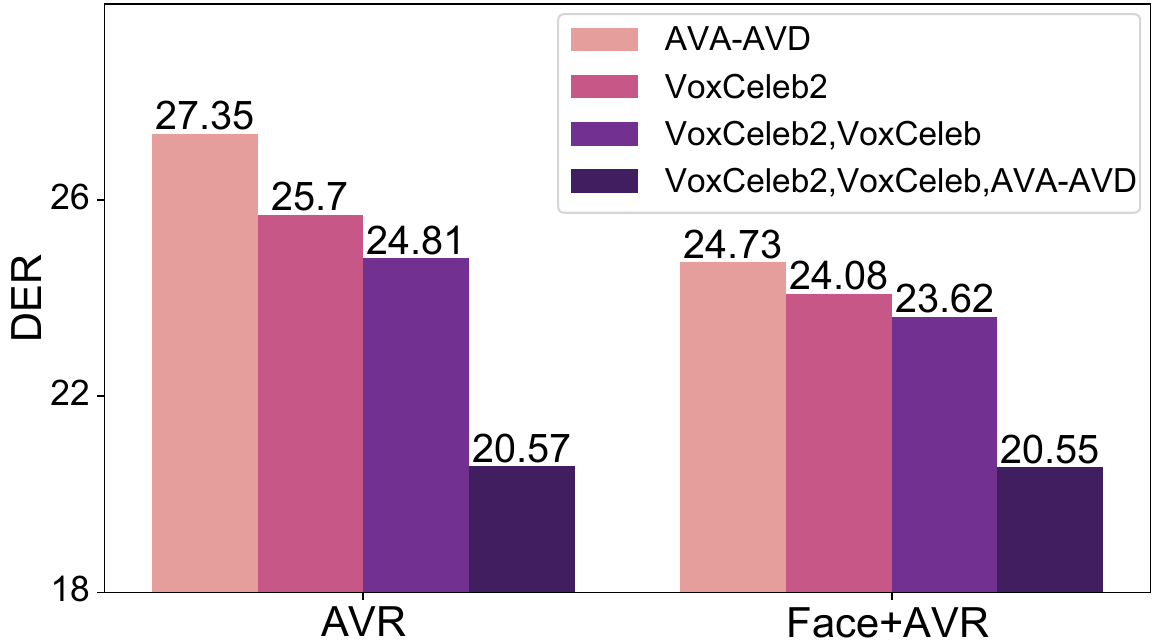}
  \caption{Diarization results of AVR-Net trained on different datasets. Although our data is relatively small, adding it to training data significantly decreases Diarization Error Rate (DER). 
  Face+AVR represents the results of AVR-Net combined with a pretrained face recognition model~\cite{deng2019arcface} using late-fusion.
  }
  \label{fig:train}
\end{figure}
\noindent{\bf Training data mixture.} VoxCeleb and VoxCeleb2 development sets do not have speaker overlap, the combination is by far the largest speaker verification dataset and thus they are usually mixed up~\cite{ryant2019second, kenny2010bayesian} for training. To evaluate the AVA-AVD train set, we train AVR-Net on different dataset combinations. The diarization results on AVA-AVD test set are shown in Figure~\ref{fig:train}. Although the scale of AVA-AVD is much smaller, the model trained only on AVA-AVD achieves a {\it comparable} result (27.35) to the model trained only on VoxCeleb2 (25.70). With the increasing amount of data, the performance can be further improved. DER drops slightly by 3.5\% when the VoxCeleb2 and VoxCeleb are mixed. Yet, adding AVA-AVD yields a larger performance gain, and the DER decreases further from 24.81 to 20.57 by a margin of 17.1\%. The main reason is the domain gap,  AVA-AVD has complex scenarios while VoxCeleb series are mainly extracted from interviews.

\noindent{\bf Fused with pretrained model.} Although the mixture of VoxCeleb series datasets is the largest dataset, the scale of face data in this combination is not comparable to large-scale face recognition datasets~\cite{cao2018vggface2,guo2016ms}. Therefore, we also use a pretrained face recognition model~\cite{deng2019arcface} and fuse the predicted similarity scores with AVR-Net to evaluate the performance again. 
For the models trained only on AVA-AVD or VoxCeleb series datasets, this late fusion strategy improves their performance. But the improvement for the models trained on three datasets is negligible. Although late-fused with a pretrained model, AVR-Net trained with AVA-AVD can still outperform other combinations without AVA-AVD, which demonstrates the effectiveness of AVA-AVD training data despite its scale is relatively small. 

\subsection{Comparisons with State-of-the-art}
\label{exp:compare}
In this section, we compare AVR-Net with state-of-the-art diarization methods on the AVA-AVD benchmark. We first keep the diarization system unchanged and replace our AVR-Net with the following SOTA verification models:

\noindent{\bf Speech+Face} are off-the-shelf face recognition model~\cite{deng2019arcface} and speaker verification model~\cite{chung2020in} which are pretrained by ourselves; 

\noindent{\bf DIMNet} is a disjoint mapping network~\cite{wen2018disjoint}, it learns to map the speech and face into a shared embedding space, where embeddings of the same person are close and otherwise far apart. It is effective both in the uni-modal and cross-modal verification~\cite{sari2021multi};

\noindent{\bf GMU} is the SOTA audio-visual person verification model~\cite{qian2021audio}, which learns to fuse audio and visual embeddings using a gated multimodal unit.

All of the above models are trained on the same datasets (VoxCeleb, VoxCeleb2, and AVA-AVD). Additionally, we also make system-level comparisons with the SOTA audio-only and audio-visual diarization systems: {\bf VBx}~\cite{landini2022bayesian} and {\bf WST}~\cite{chung2019said}.
\begin{table}[t]
\centering
\small
\caption{Comparisons with state-of-the-art methods. Our system outperforms SOTA methods on AVA-AVD.}
\begin{subtable}{\linewidth}
\centering
\caption{Comparison between AVR-Net and SOTA speaker verification models. Avg SPKE and Avg DER are average results with different missing rates.}
\begin{tabular}{lcccc}
\hline
\multicolumn{1}{l}{} & SPKE & DER & Avg SPKE  & Avg DER\\
\hline
\multirow{1}{*}{Speech\cite{chung2020in}+Face\cite{deng2019arcface}}
&21.48&24.40&22.98&25.90 \\
\multirow{1}{*}{DIMNet\cite{wen2018disjoint}}
&19.77&22.69&21.06&23.98 \\
\multirow{1}{*}{GMU\cite{qian2021audio}}
&19.10&22.02&20.10&23.02 \\
\multirow{1}{*}{Ours}
&{\bf 17.65}&{\bf 20.57}&{\bf 19.45}&{\bf 22.37} \\
\hline
\end{tabular}\\

\label{subtab:baseline}
\end{subtable}

\begin{subtable}{\linewidth}
\centering
\caption{System-level comparisons. We also ablate the prior steps (voice activity detection (VAD), audio-visual correspondence (AVC)) in our system. E denotes a simple energy-based VAD method.}
\begin{tabular}{lcccccc}
\hline
\multicolumn{1}{l}{} & VAD & AVC & MS & FA & SPKE & DER\\
\hline
VBx\cite{landini2022bayesian}&E&NA&8.71&44.55&35.27&88.53 \\
(ResNet34)&GT&NA&2.92&0.0&35.14&38.06 \\
\hline
\multirow{2}{*}{WST\cite{chung2019said}}
&E&SyncNet\cite{Chung16a}&11.63&40.63&36.15&88.41 \\
&GT&SyncNet\cite{Chung16a}&3.08&0.0&38.95&42.04 \\
\hline
\multirow{4}{*}{Ours}
&E&TalkNet\cite{tao2021someone}&8.71&44.55&20.09&73.34 \\
&E&GT&8.71&44.55&16.88&70.13 \\
&GT&TalkNet\cite{tao2021someone}&2.92&0.0&21.25&24.18 \\
&GT&GT&2.92&0.0&17.65&20.57 \\
\hline
\end{tabular}\\
\label{subtab:system}
\end{subtable}
\end{table}

\noindent{\bf Comparison with state-of-the-art.} Table~\ref{subtab:baseline} shows that our AVR-Net outperforms all of the counterparts. The pretrained baseline, Speech+Face, achieves a DER of 24.18. Our AVR-Net yields the {\it best} result 20.57 because it learns to compute the similarity of two audio-visual pairs, while the pretrained model only employs linear similarity metric and suboptimal late-fusion strategy. Also, the cross-modal biometrics are ignored by this pretrained baseline. DIMNet maps audio and visual inputs to the same embedding space by sharing a classifier, but it is an extension of the pretrained baseline and does not explore the cross-modal links between face and voice from different speakers. AVR-Net outperforms GMU by 6.6\%. Although GMU can learn cross-modal attention to fuse the audio and visual embeddings dynamically, it is a single-stream model which is harder to be optimized on diarization dataset.

To study the system performance and the effects of each component in prior steps, we further make system-level comparisons in Table.~\ref{subtab:system}. We report diarization error rates with both ground truth and predicted voice activity detection (VAD) and active speaker detection (ASD) results. First, our system has a large performance drop when the VAD changes from ground truth to predicted results, which is in line with the analysis of challenges in Sec.~\ref{exp:challenge}. Another important prior step in our system is the ASD, we use an off-the-shelf active speaker detection method TalkNet~\cite{tao2021someone}, which is trained only on the AVA-Active Speaker Detection dataset. The change of ASD from ground truth to TalkNet brings an absolute performance drop of around 3.5. Nevertheless, our system outperforms WST using both ground truth VAD (24.18 {\it vs.} 42.04) and predicted ones (73.34 {\it vs.} 88.41). Compared with the audio-only VBx, our system demonstrates improvements of leveraging visual information. To make a fair comparison, here the backbone of VBx is switched to ResNet34.

\begin{figure}[ht]
  \centering
  \includegraphics[width=0.35\textwidth]{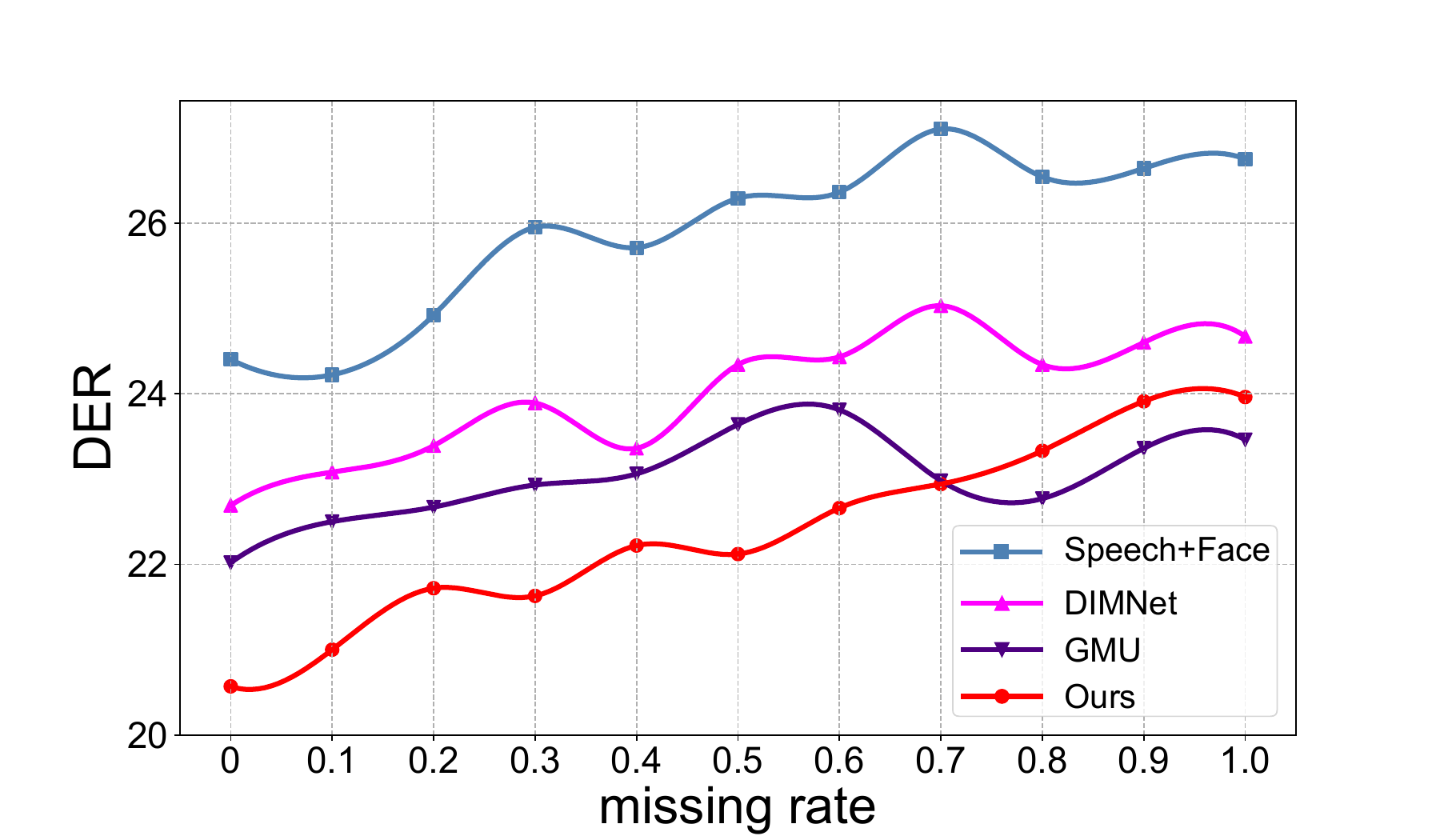}
  \caption{Diarization Error Rate (DER) under different face missing rates. 
  }
  \label{fig:missing}
\end{figure}

\begin{figure}[ht]
  \centering  
  \includegraphics[width=0.37\textwidth]{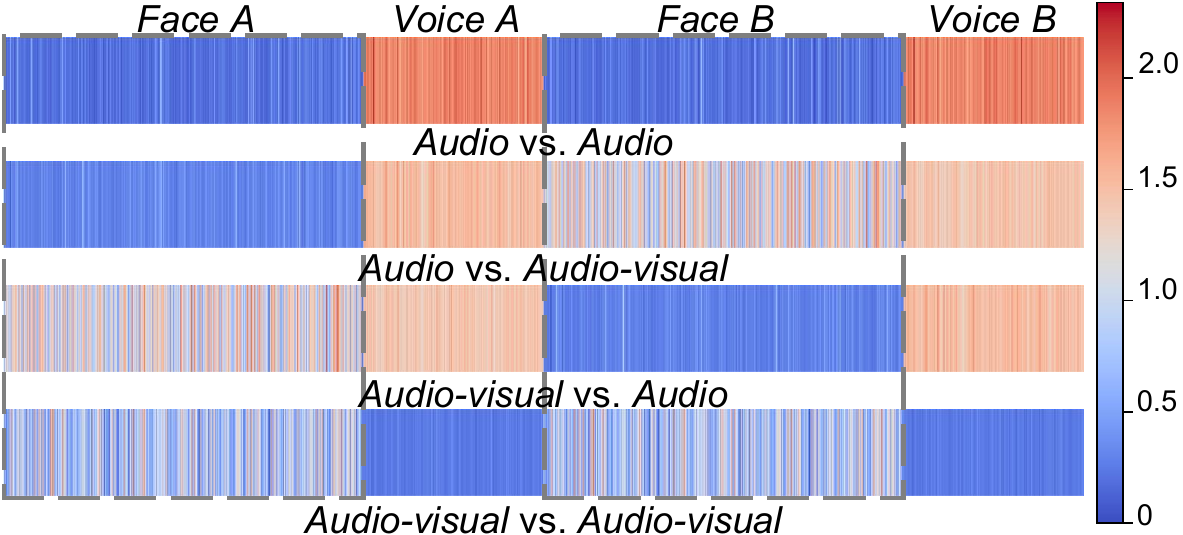}
  \caption{Modality masks learned by AVR-Net. Red represents larger weights while blue represents smaller ones. 
  Our modality mask can dynamically attend to different channels depending on the visibility.
  }
  \label{fig:mask}
\end{figure}

\noindent{\bf Robustness against increasing missing face.} AVA-AVD has a large number of off-screen speakers, the original missing faces in AVA-AVD account for 50\%. To investigate the robustness of models against missing faces, we further synthesize {\it additional} missing faces with a certain probability (missing rate) and evaluate the diarization performance. Figure~\ref{fig:missing} shows that our AVR-Net performs best when the missing rate is less than 80\%. When the missing rate increases further, GMU is slightly better than ours, but both ours and GMU outperform other methods. The average values are summarized in Table~\ref{subtab:baseline}. Although our model cannot outperform GMU under high missing rates (>=80\%), the average SPKE 19.45 and average DER 22.37 are still the lowest, showing the robustness of AVR-Net against missing faces.

\begin{figure*}[t]
  \centering
  \begin{subfigure}{0.75\linewidth}
  \centering
  \includegraphics[width=\linewidth]{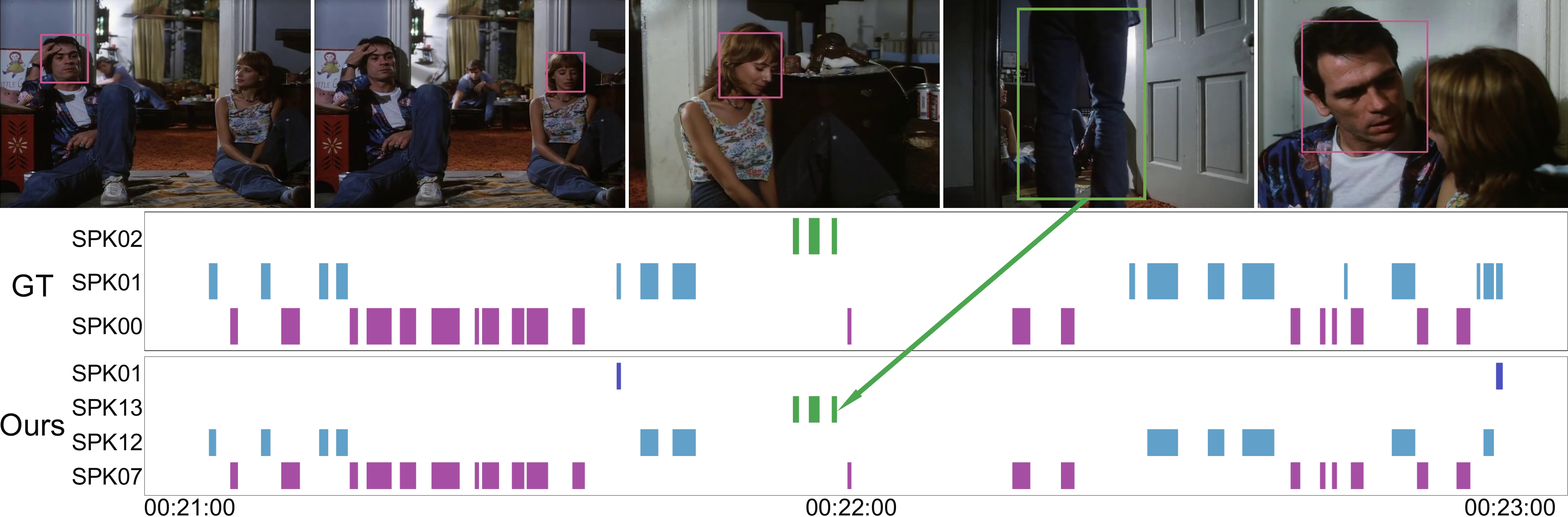}
  \caption{Results for video oD\_wxyTHJ2I from 21 min to 23 min. The person highlighted in the green box is speaking but his face is invisible. Our system can successfully diarize this invisible speaker.}
  \label{fig:qual1}
  \end{subfigure}
  \begin{subfigure}{0.75\linewidth}
  \centering
  \includegraphics[width=\linewidth]{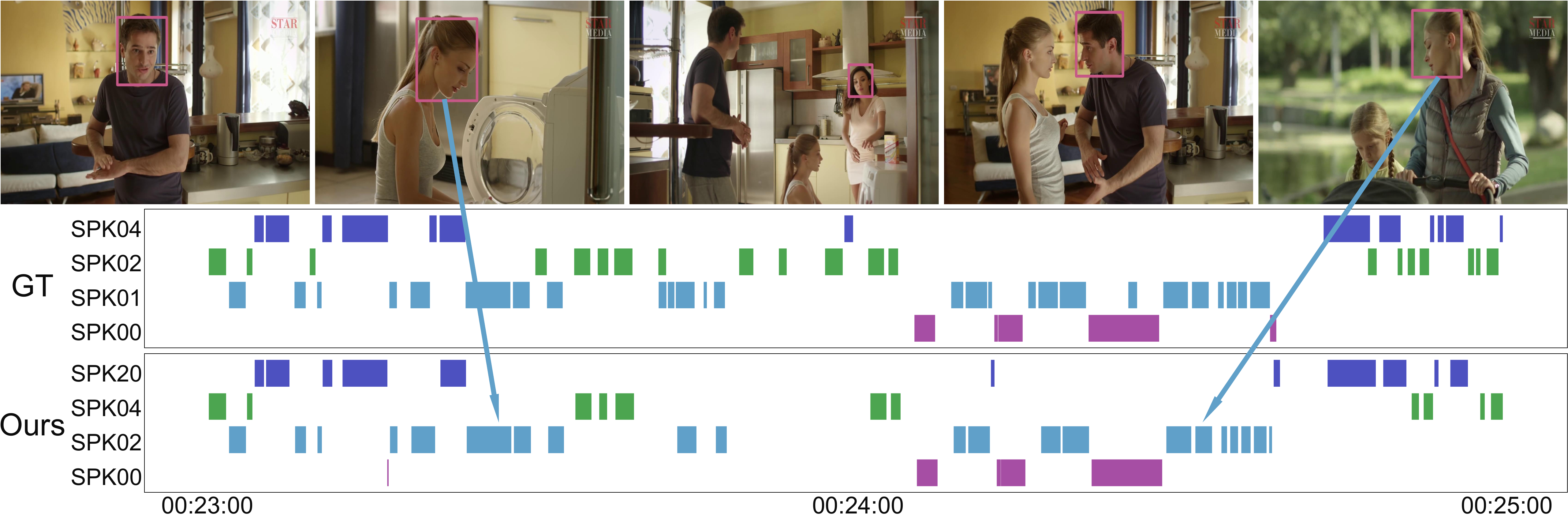}
  \caption{Results for video zR725veL-DI from 23 min to 25 min. There are scene changes in this short clip, our system is able to diarize SPK02 across different scenes.}
  \label{fig:qual2}
  \end{subfigure}
  \caption{Qualitative results of our system. Persons highlighted in pink boxes are active speakers. The wrongly identified segments shorter than 1 second are ignored.} 
  \label{fig:qual}
\end{figure*}

\subsection{Qualitative Results}
\label{exp:qual}
Figure~\ref{fig:qual} demonstrates the qualitative results of our system. Compared with ground truth, our system has an overall good diarization performance in complex scenarios. As shown in Figure~\ref{fig:qual1}, the speaker in the green box is invisible, yet our system can successfully recognize all of his utterances. In addition, our AVA-AVD has frequent scene changes. Although there are changes in the background and acoustic environment, our system can still diarize the female speaker (SPK02) across scenes. 

Our system adopts unsupervised clustering and the number of speakers is {\it unknown}. Thus, our system predicts several wrong speakers (SPKE). The wrong segments which is less than 1 s are ignored.

\subsection{Analysis of modality masks}
\label{exp:analysis}
To further interpret and understand the learnable modality masks, we visualize the weights in Figure~\ref{fig:mask}. Our AVR-Net digests two audio-visual pairs and each row represents the learned mask weights for each comparison case. Each vertical line in the row represents the activation weight of this channel during inference. Red denotes a higher weight, while blue denotes a lower one.

The activation weights reveal that: (1) when both of the faces are invisible, the network will only rely on audio inputs, different channels of audio inputs are almost {\it equally important}; (2) if only one face is visible, the network will not simply skip this face. The only face can still help compute the similarity between two candidate pairs. This cross-modal relationship between faces and voices is also well studied in the cross-modal verification tasks. In this case, our AVR-Net not only can compute similarities based on audio streams but also utilizes the {\it cross-modal links}. This mechanism makes our system fully utilize the existing information and thus more robust against missing faces that are quite common in in-the-wild videos; (3) when both speakers have visible faces, only part of the channels in face feature maps are activated. It means that (i) faces are more discriminative than audio signals for our task because the face backbones are pretrained on a much larger scale dataset; (ii) the activation of faces are {\it sparse}, which means that some attributes of a face image are more discriminative than others.

\section{Conclusion}
In this work, we introduce a new challenging audio-visual diarization dataset AVA-AVD. We experimentally show that both the train and test sets in AVA-AVD are useful. Also, we propose a simple yet effective audio-visual relation network, which learns to predict similarities using modality masks. Experiments demonstrate that our method outperforms state-of-the-art audio-visual diarization systems and is more robust to missing faces.

\section{Acknowledgement}
This project is supported by the DSO National Laboratories. The computational work for this article was partially performed on resources of the National Supercomputing Centre, Singapore.

\newpage
\appendix
\section{Appendix}

\subsection{Train/val/test split.} We split AVA-AVD dataset into train, validation, and test subsets. Current audio-visual diarization datasets~\cite{chung2020spot,brown2021face} do not hold out training data because: (1) the number of identities is limited; (2) diarization dataset only provides video-level identity labels. However, considering that AVA-AVD has a large number of identities, it could potentially benefit training audio-visual {\it speaker verification} models and {\it end-to-end diarization} models. Thus, we still partition train set and the necessity of training data is discussed in Sec.~\ref{exp:train}. Since each 15-minute video is divided into three 5-minute clips, we do not split the clips from the same movie into different subsets to keep the disjoint property across subsets. Therefore, our split can reflect models' generalization ability in unseen scenarios.
In addition, we also keep the test split consistent with the validation set in AVA-Active Speaker. The main advantage is that an active speaker detection model pretrained on AVA can avoid speaker leakage issues and can directly be used for inference. 

\subsection{Ablation Study}
\label{apex:ablations}

\noindent{\bf Learning objective.} Training AVR-Net is not straightforward because AVA-AVD only has video-level labels. Two solutions are discussed in Sec.~\ref{AVR-Net}. Table~\ref{apex:ablations:stream} compares the Siamese network~\cite{koch2015siamese} with single-stream architecture. It shows that single-stream network has a much higher DER than AVR-Net. AVR-Net can regress the similarity of two speakers, this objective is not as strict as the classification supervision used in the single-stream network. Our architecture is well-suited for the large-scale diarization dataset with only video-level identity labels.

\noindent{\bf Input features.} Recent face clustering~\cite{tapaswi2019video}, multi-modal clustering~\cite{hu2017learning,jiang2019dm2c}, and verification~\cite{qian2021audio} methods learn the threshold or unified representation based on the embedding vectors extracted by pretrained models. Hence, we also study the influence of inputs on our AVR-Net. We have three input options: (1) raw face and audio stream; (2) embeddings extracted by pretrained models; (3) feature maps encoded by pretrained backbones.

Table~\ref{apex:ablations:feat} shows that the AVR-Net using raw input has a much higher DER of 28.37. Although the embedding vectors are discriminative enough and can achieve 21.00, our AVR-Net architecture yields a better result because AVR-Net encodes informative features which are more discriminative and helpful for computing cross-modal links and comparing speakers.

The modality mask in AVR-Net is multiplied by the input features. Table~\ref{apex:ablations:mask} studies other alternatives, i.e. concatenating masks with the input features. Moreover, the modality masks are passed into the relation module, they are part of the input signals. Therefore, we also study the alternative training manners. Results in Table.~\ref{apex:ablations:alter} demonstrate that: (1) multiplied modality mask is more effective than concatenation; (2) alternative training cannot boost the performance. Our AVR-Net updates modality masks simultaneously during training, it is simple yet effective.

In addition to learnable masks, binary modality masks (i.e. 1 if visible, 0 vice versa) can be manually assigned to the features depending on the visibility. Table~\ref{apex:ablations:mask} summarizes the comparisons between hand-crafted masks and learnable ones. Using no modality masks, hand-crafted masks, and partially learnable masks have similar performance. While learnable modality masks are optimized during training and can learn to activate the most informative or discriminative features, thus it is consistently better.

\noindent{\bf Missing probability.} During training, to synthesize the cases where faces are invisible, we apply a missing face augmentation. Each speaker's face has a certain probability to be removed. Table~\ref{apex:ablations:mr} studies the missing probability. The missing probability of 0 achieves comparable results with 0.5. In this case, the missing face rate is close to the test set. However, since there is no synthesized missing faces, the model cannot learn the relations for {\it A} vs. {\it A}, {\it A} vs. {\it AV}, and {\it AV} vs. {\it A} effectively on VoxCeleb and VoxCeleb2. Another extreme case is 100\% missing probability, where no faces are passed into the AVR-Net, thus the performance deteriorates a lot and DER is only 53.40. Among all of the values, 0.5 has the best result since it achieves a balance between visible and invisible speakers.

\subsection{Participants}
\label{apx:irb}
The video source of AVA-AVD is an existing public video dataset. We do not record any new videos and thus do not recruit any participants for data collection. Thus, we could waive the participant-related requirements from Institutional Review Board (IRB) protocols. In total, we recruit 10 annotators and 2 auditors. The only requirement for these human annotators is that they must be able to identify and localize a person by leveraging both visual and auditory cues.

\subsection{Annotation Tool}
\label{apx:tools}
\begin{figure*}[h]
  \centering
  \includegraphics[width=0.9\linewidth]{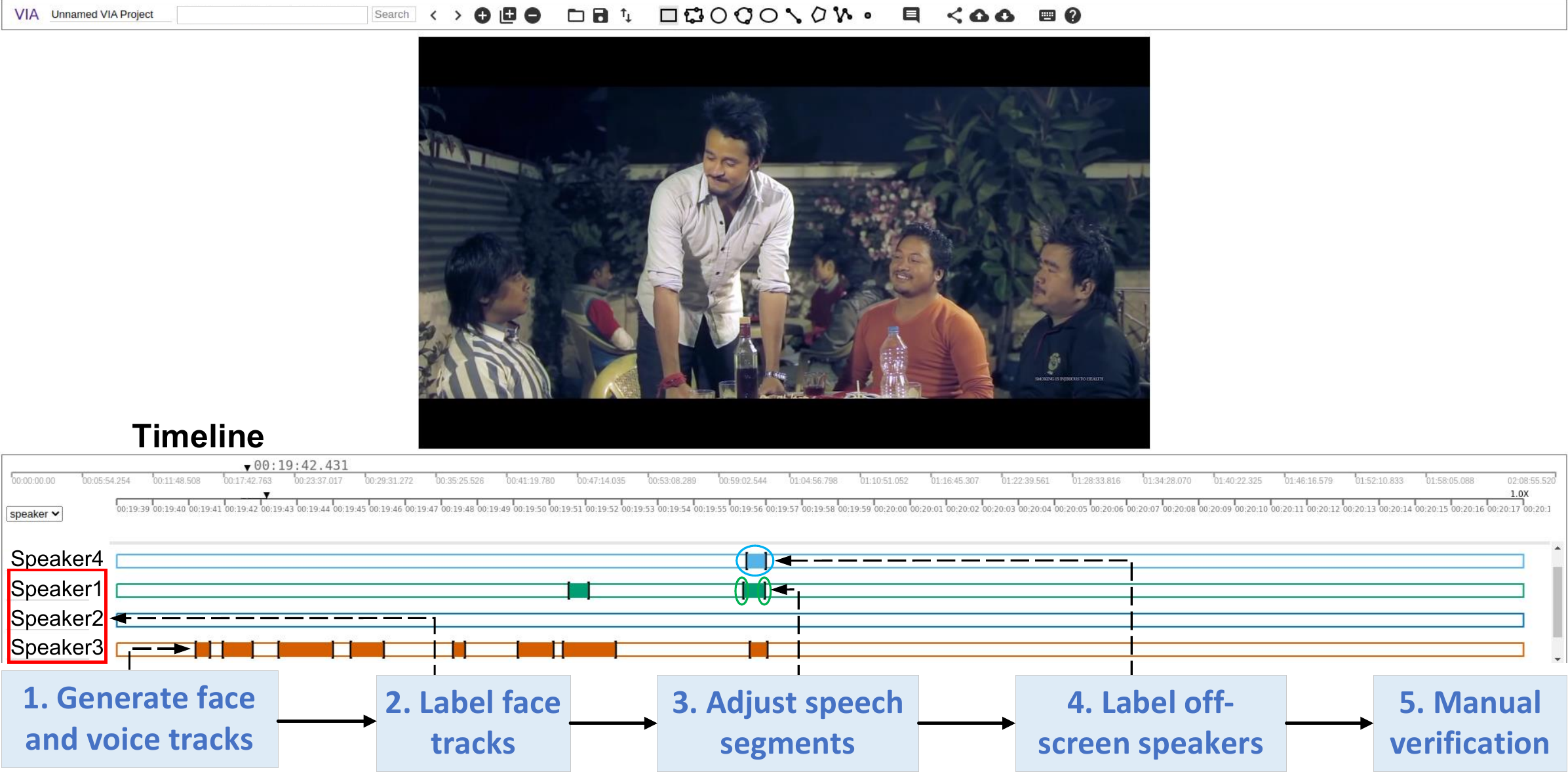}
  \caption[]{Annotation tool and process. Our annotation tool is based on VIA. Details of the annotation steps can be found in Sec.~\ref{annotation}.} 
  \label{fig:annot}
\end{figure*}

\noindent Our annotation tool is developed based on an open-source project VIA\footnote{https://gitlab.com/vgg/via}. Its GUI is visualized in Figure~\ref{fig:annot}.


\begin{table*}[t]
\centering
\begin{subtable}{0.264\linewidth}
\centering
\begin{tabular}{lcc}
\hline
\multicolumn{1}{l}{} & SPKE & DER\\
\hline
\multirow{1}{*}{Single}
&22.17 &25.09 \\
\multirow{1}{*}{Siamese}
&{\bf 17.65}&{\bf 20.57} \\
\hline
\\
\\
\end{tabular}
\caption{Model architecture.}
\label{apex:ablations:stream}
\end{subtable}\begin{subtable}{0.264\linewidth}
\centering
\begin{tabular}{lcc}
\hline
\multicolumn{1}{l}{} & SPKE & DER\\
\hline

\multirow{1}{*}{Raw}
&25.45&28.37 \\

\multirow{1}{*}{Embed}
&18.08&21.00 \\
\multirow{1}{*}{Feat}
&{\bf 17.65}&{\bf 20.57} \\
\hline
\\
\end{tabular}
\caption{Input features.}
\label{apex:ablations:feat}
\end{subtable}\begin{subtable}{0.312\linewidth}
\centering
\begin{tabular}{lcc}
\hline
\multicolumn{1}{l}{}& SPKE & DER\\
\hline

\multirow{1}{*}{Concat}
&19.37&22.29 \\

\multirow{1}{*}{Alter Concat}
&19.69&21.98 \\
\multirow{1}{*}{Mask}
&{\bf 17.65}&{\bf 20.57} \\

\multirow{1}{*}{Alter Mask}
&17.87&20.79 \\
\hline
\end{tabular}
\caption{Alternative training.}
\label{apex:ablations:alter}
\end{subtable}

\begin{subtable}{0.38\linewidth}
\centering
\begin{tabular}{lcc}
\hline
\multicolumn{1}{l}{}& SPKE & DER\\
\hline

\multirow{1}{*}{None}
&18.37&21.29\\

\multirow{1}{*}{Hand-crafted Mask}
&18.44&21.36\\

\multirow{1}{*}{Partially Learnable Mask}
&18.54&21.46\\
\multirow{1}{*}{Learnable Mask}
&{\bf 17.65}&{\bf 20.57}\\
\hline
\\
\\
\end{tabular}
\caption{Learnable modality masks.}
\label{apex:ablations:mask}
\end{subtable}\begin{subtable}{0.38\linewidth}
\centering
\begin{tabular}{lcc}
\hline
\multicolumn{1}{l}{prob} & SPKE & DER\\
\hline
\multirow{1}{*}{0.0}
&18.02 &20.94 \\
\multirow{1}{*}{0.1}
&19.23 &22.15 \\
\multirow{1}{*}{0.3}
&19.12&22.04 \\
\multirow{1}{*}{0.5}
&{\bf 17.65}&{\bf 20.57} \\
\multirow{1}{*}{0.7}
&21.01&23.93\\
\multirow{1}{*}{1.0}
&50.48&53.40\\
\hline
\end{tabular}
\caption{Missing face probability.
}
\label{apex:ablations:mr}
\end{subtable}
\caption{AVR-Net ablation experiments. We report diarization performance of different alternatives.
}
\label{tab:task}
\end{table*}

\newpage
\bibliographystyle{ACM-Reference-Format}
\bibliography{ref}

\end{document}